# Generative adversarial networks (GAN) based efficient sampling of chemical space for inverse design of inorganic materials

Yabo Dan[1]  Yong Zhao[2]  Xiang Li[1] Shaobo Li[1,4] Ming Hu[3] and Jianjun Hu[1,2,*]

[1]School of Mechanical Engineering, Guizhou university, Guiyang 550025, China;

[2]Department of Computer Science and Engineering, University of South Carolina, Columbia 29201，USA;

[3]Department of Mechanical Engineering, University of South Carolina, Columbia 29201，USA;

[4]Key laboratory of advanced manufacturing technology, Ministry of education, Guiyang 550025, China;

Correspondence:  Jianjun Hu (Jianjunh@cse.sc.edu)

**Abstract**:

A major challenge in materials design is how to efficiently search the vast chemical design space to find the materials with desired properties. One effective strategy is to develop sampling algorithms that can exploit both explicit chemical knowledge and implicit composition rules embodied in the large materials database. Here, we propose a generative machine learning model (MatGAN) based on a generative adversarial network (GAN) for efficient generation of new hypothetical inorganic materials. Trained with materials from the ICSD database, our GAN model can generate hypothetical materials not existing in the training dataset, reaching a novelty of 92.53% when generating 2 million samples. The percentage of chemically valid (charge neutral and electronegativity balanced) samples out of all generated ones reaches 84.5% by our GAN when trained with materials from ICSD even though no such chemical rules are explicitly enforced in our GAN model, indicating its capability to learn implicit chemical composition rules. Our algorithm could be used to speed up inverse design or computational screening of inorganic materials.

# 1. Introduction

Discovering new inorganic materials such as solid electrolytes for lithium-ion batteries is fundamental to many industrial applications. While recent years have observed tremendous efforts on rational materials design, progress has been limited due to the challenge to find new materials that meet diverse technical and economic constraints. From the computational perspective, brute-force molecular simulations or first-principles methods are computationally too expensive for large-scale screening of the vast chemical space. A recent effort (1) to quantify the magnitude of the compositional space for multi-component inorganic materials showed that even after the application of chemical filters such as charge neutrality or electronegativity balance, the space for four-component/element materials exceeds $10^{10}$ combinations and the five-component/element space exceeds $10^{13}$ combinations. Indeed, a machine learning based model has been applied to screen billions of hypothetical materials to identify promising high ion-conductors (2). Considering the huge space of doped materials with different mixing ratios of elements and many applications such as high-temperature superconductors, where six to seven component materials are common, the number of potential materials is immense. Such combinatorial explosion calls for the need for more effective sampling approaches to search the chemical design space that employ existing explicit chemical and physical knowledge and also implicit elemental composition knowledge embodied within known synthesized materials. To gain more efficient search, a variety of explicit chemical rules for assessing the feasibility of a given stoichiometry and the likelihood of particular crystal arrangements have been used in computational screening such as the Pauling's rules (charge neutrality), electronegativity balance, the radius ratio rules (3), Pettifor maps (4) and etc. However, such approaches still fail to capture enough implicit chemical rules to achieve efficient chemical design space sampling.

Recently, generative machine learning models such as autoencoders (AE) and its variants (VAE, AAE), RNNs, Generative Adversarial Networks (GANs) have been successfully applied to inverse design of organic materials (5) (6) (7) (8). These algorithms mainly exploit the sequential or graph representations of organic materials to learn the composition rules of the building blocks for generating valid and novel hypothetical materials. Given a large set of samples, a GAN is capable of learning complicated hidden rules that generate the training data, and then applies these learned rules to create new samples with target properties. When applied to inverse design, GANs have demonstrated their power in efficient sampling of design space(5, 9), more efficient than other sampling approaches such as random sampling(10), Monte Carlo sampling, and other heuristic sampling (such as genetic algorithms (11)). However, due to the radical difference in building blocks and their composition rules, such generative machine learning models have not been applied to the generation of inorganic materials so far to the best of our knowledge. Recently, variational autoencoders (11, 12) have been proposed to generate hypothetical crystal structures of inorganic materials. However, these methods are either limited to generate new structures of a given material system such as the V-O system (11) or cannot generate molecules that are physically stable (12).

In this paper, we propose the first generative adversarial network model for efficient sampling of inorganic materials design space by generating hypothetical inorganic materials. Trained with materials from inorganic materials databases such as OQMD(13), Materials Project(14), and ICSD, our GAN models are able to learn the implicit chemical compositional rules from the known materials to generate hypothetical but chemically sound compounds. Without explicitly specifying the chemical rules, our GANs trained with all charge-neutral and electronegativity-balanced samples of the ICSD subset can generate hypothetical materials with 84.5% reproducing the charge-neutrality and balanced electronegativity. The analysis shows that our generative GAN can achieve much higher efficiency in sampling the chemical composition space of inorganic materials than the exhaustive enumeration approach.

## 2. Results
### 2.1 Representation of inorganic materials

Through simple statistical calculation of the materials in the OQMD dataset (15), 85 elements are found and each element usually has less than 8 atoms in any specific compound/formula. We then represent each material as a sparse matrix $T \in R^{d \times s} (d=8, s=85)$ with 0/1 cell values. Each column represents one of the 85 elements while the column vector is a one-hot encoding of the number of atoms of that specific element.

### 2.2 The GAN generation model

Generative models can be built on several machine learning algorithms such as variational autoencoder (VAE), generative adversarial networks (GAN), Reinforcement learning (RL), Recurrent Neural networks (RNN), and their hybrids [5]. Different from other generative models [16, 17], GANs do not directly use the discrepancy of the data and model distributions to train the generator. Instead, it uses an adversarial training approach: it first trains a discriminator to differentiate real samples from faked samples, which then guides the training of the generator to reduce this difference. These two training processes are alternatively repeated. Their arm race will lead to high performance of both the generator and the discriminator.

Our generative ML model for inorganic materials (MatGAN) is based on the GAN scheme as shown in Fig.1.

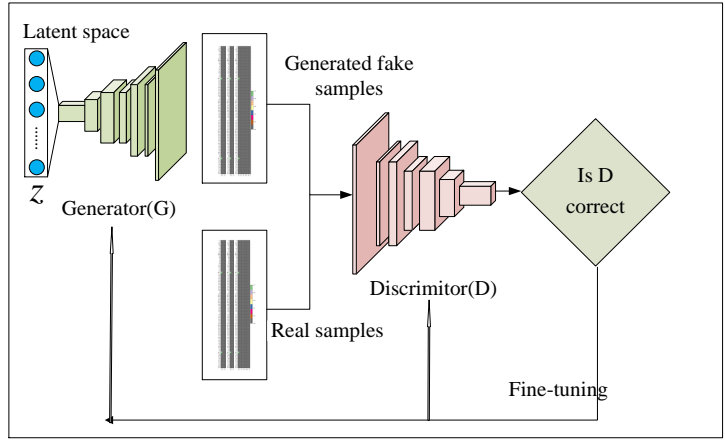

Fig.1 Architecture of MatGAN for inorganic materials. It is composed of a generator, which maps random vectors into generated samples and a discriminator, which tries to differentiate real materials and generated ones. Detailed configuration parameters are listed in supplementary Table S1 and S2 and supplementary Fig. S1.

We choose the 8×85 matrix representation of materials samples to build the GAN model. We found the integer representation of materials greatly facilities the GAN training. In our GAN model, both the discriminator (D) and the generator (G) are modeled as a deep neural network. The generator is composed of 1 fully connected layer and seven deconvolution layers. The discriminator is composed of seven convolution layers followed by a fully connected layer. Each of the convolution and deconvolution layer comes with a batch normalization layer. The output layer of the generator uses the Sigmoid function as the activation function while all other batch normalization layers use the ReLu as the activation function. The detailed network configuration is shown in supplementary Table S1 and S2. In order to avoid the gradient vanishing issue of standard GANs, we adopt the Wasserstein GAN(18), which replaces the JS divergence distance with the Wasserstein distance. The GAN model will be trained using the Wasserstein GAN approach by minimizing both the generator loss and discriminator loss, which are defined as

$$Loss_G = -\mathrm{E}_{x \sim P_g}\left[f_w(x)\right] \qquad (1)$$

$$Loss_D = \mathrm{E}_{x \sim P_g}\left[f_w(x)\right] - \mathrm{E}_{x \sim P_r}\left[f_w(x)\right] \qquad (2)$$

where, $P_g, P_r$ is the distributions of generated samples and real samples; $f_w(x)$ is the discriminant network. Equations (1) and (2) are used to guide the training process. The smaller the $Loss_D$, the smaller the Wasserstein distance between the generated samples and the real samples and the better the GAN is trained.

## 2.3 Variational Autoencoder for Evaluating GAN performance

During our GAN generation experiments for OQMD dataset, we found that it sometimes has difficulty to generate a specific category of materials. This may be caused by the limited samples to learn the required composition rules to generate those samples. To investigate this issue, we built an autoencoder (AE)[19] model as shown in Fig.2. The autoencoder is composed of an encoder with seven convolutional layers followed by a fully connected layer and a decoder composed of a fully connected layer followed by seven deconvolution layers. After each of the convolution and deconvolution layer, there is a batch normalization layer used to speed up training and reduce the influence of initial network weights (20). The ReLu is used as the activation function for all the batch normalization layers. The Sigmoid function is used as the activation function for the decoder's output layer. The detailed configuration parameters are listed in supplementary Table S3. The autoencoder are trained with 291,840 inorganic materials selected from the OQMD database. In order to ensure the overlap between the original input matrix T and the matrix reconstructed by the decoder as much as possible, we adopt the negative dice coefficient (21) commonly used in medical image semantic segmentation as the loss function of AE. The AE model is then trained using the back-propagation algorithm. The loss function is shown in the following equation:

$$Loss_{AE} = -Dice = -\frac{2|A \cap B|}{|A|+|B|} \approx -\frac{2 \times A \bullet B}{Sum(A)+Sum(B)} \quad (3)$$

where $A \cap B$ denotes the common elements of $A$ and $B$, $|g|$ represents the number of elements in a matrix, • denotes dot product, $Sum(g)$ is the sum of all matrix elements. Dice coefficient essentially measures the overlap of two matrix samples, with values ranging from 0 to 1 with 1 indicating perfect overlap (22).

The decoder module of the AE model shares the same architecture of the generator in our MatGAN model. Our hypothesis is that if the trained AE model cannot decode a specific material, it is unlikely our GAN model can generate it. By screening out the non-decodable materials out of the OQMD database using the AE, we may obtain a deeper understanding of the limitations of our GAN models.

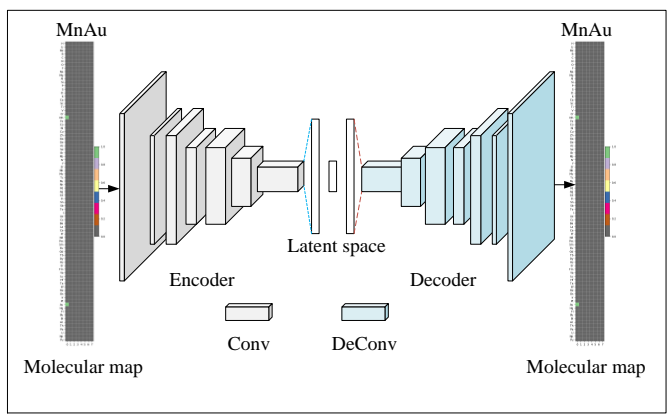

Fig. 2 Architecture of Autoencoder. Detail configuration parameters are shown in supplementary Table S3 and supplementary Fig S2.

## 2.4 The model performance of GANs

**Efficient sampling of the inorganic chemical space by the GANs:** We trained our GANs according to the procedures as detailed in Methods. For all these GANs including GAN-OQMD, GAN-MP, GAN-ICSD, we then generated 2 million hypothetical materials using each of these generators and evaluate their validity, uniqueness, and novelty.

**Mapping inorganic materials design space**

Out of the 2 million samples generated by the GAN-ICSD, we filter out all samples that do not satisfy charge neutrality and balanced electronegativity leading to 1.69 million generated samples. To visualize how the generated ones are distributed compared to the training datasets from ICSD, we applied T-sne dimension reduction technique (23) to reduce the dimension of the matrix representations of the samples from the generated set, the training set, and the leave-out validation set. The distribution of the generated samples versus the training and validation set are shown in Fig. 3. It is observed that the training samples from ICSD occupy only a very small portion of the whole space. The GAN-ICSD, however, has been able to generate potentially interesting hypothetical materials that fill the design space, which may significantly expand the range of the ICSD database.

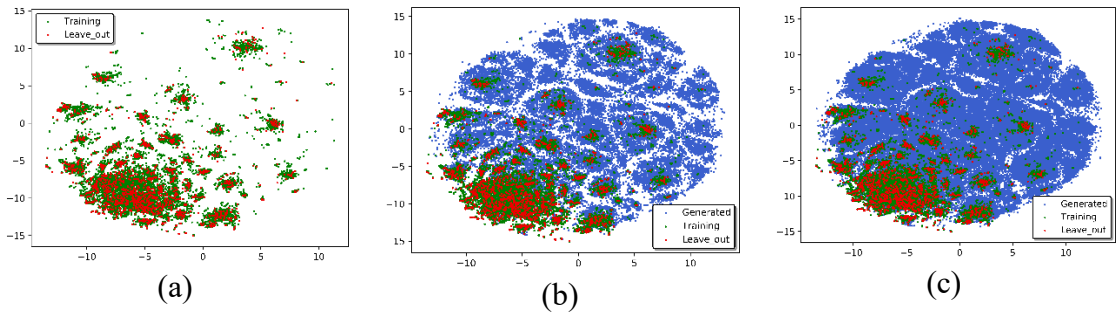

(a)　　　　　　　　　　(b)　　　　　　　　　　(c)

Fig. 3. Inorganic materials space composed of existing ICSD materials and hypothetical materials generated by GAN-ICSD. The two axes correspond to the two dimensions after t-sne based dimension reduction. The ICSD materials only occupies a tiny portion of the chemical space of inorganic materials. (a) Training samples (green dots) and leave-out validation samples (red dots) from ICSD; (b) 50,000 generated samples (blue dots) together with training and leave-out samples (c) 200,000 generated samples together with training and leave-out samples.

**Validity check**: charge neutrality and electronegativity balance are two fundamental chemical rules of crystals. It is thus interesting to check how the generated samples from our GAN models satisfy these rules without explicit enforcement of such rules during model training. To do this, we adopt the charge-neutrality and electronegativity check procedure as proposed in Ref. (1) to calculate the percentages of samples that obey these rules within the training and generated sets of all 4 databases. The results are shown in Fig.4. First, we found that the percentages of the validly generated samples are very close to those of the training set. For OQMD, when the training set has 55.8% charge-neutral samples, the generated set has 56.1%. For MP and ICSD, the percentage of generated charge-neutral samples (84.8% and 80.3%) are also close to those of the training sets (83.5% and 84.4%). Similar observations are found for electronegativity check. It is impressive that when we ensure all training samples in the ICSD_filter are charge-neutral and electronegativity balanced, up to 92.1% and 84.5% of the generated samples satisfy the two chemical rules, respectively, despite that no such rules are explicitly modeled or enforced in our GAN training models. To demonstrate the significance of this high percentage of chemically valid candidates, we compare our results to the exhaustive enumeration approach in (Ref (1) Table 1). The percentage of all binary/ternary/quaternary samples that satisfy both charge neutrality and electronegativity is 0.78% with exhaustive enumeration compared to our 62.24%, which corresponds to 77 times of enrichment in terms of sampling efficiency. This strongly indicates that our GAN models have successfully learned implicit chemical rules for generating chemically valid hypothetical materials.

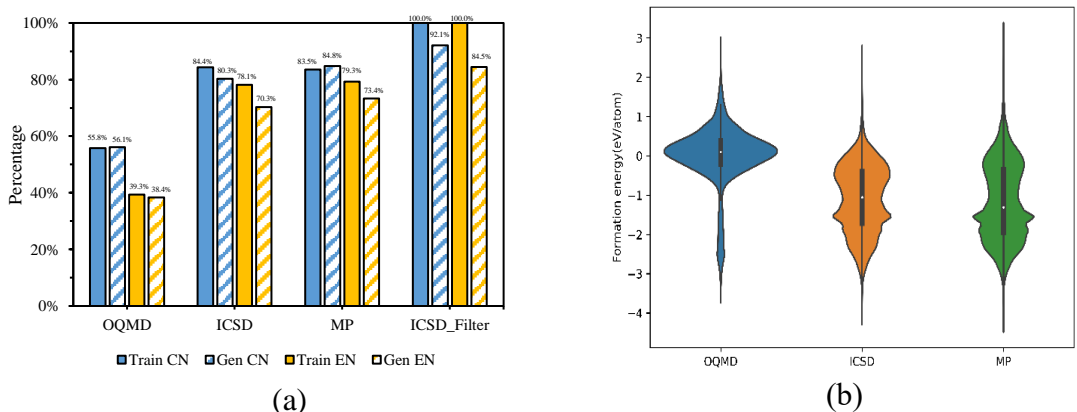

(a) (b)

Fig.4 Evaluation of the validity of generated materials. (a) the percentages of charge-neutral (CN) and electronegativity-balanced (EN) samples of the generated samples are very close to those of the training sets for all four datasets. Train/gen CN: percentage of training/generated samples that satisfy charge neutrality; Train/gen EN: percentage of samples that satisfy balanced electronegativity. (b) Formation energy distribution of the Li-containing compounds generated by three GANs. Both GAN-ICSD and GAN-MP can generate a large percentage of hypothetical materials with low ($< 0$) formation energy.

**Formation energy distribution of generated materials**: another way to evaluate the quality of generated hypothetical materials is to check their stability, which can be measured by their formation energy (24). To do this, first, GAN-OQMD, GAN-ICSD, and GAN-MP were used to generate 2 million materials candidates each. Then, we selected all the materials with lithium element and then filter out all those materials that do not satisfy charge neutrality and balanced electronegativity. Finally, we obtained 15591, 137948, and 281320 lithium-containing compounds, respectively, from GAN-OQMD, GAN-ICSD, and GAN-MP. We then downloaded the formation energy prediction machine learning model (ElemNet) developed by Jha et al. (24) and then used it to predict the formation energies of all these hypothetical materials. Fig. 4(b) shows that the formation energy of these generated materials are mostly less than 0, especially for those generated by GAN-ICSD and GAN-MP, which are trained with more chemically valid samples. Also, much higher percentage of generated samples by the GAN-OQMD are found to have higher formation energy scores in the figure, which is due to the fact that 68.48% training samples of OQMD have formation energies larger than 0.

**Uniqueness check**:

To check the uniqueness of the generated samples, we calculate the percentages of the number of unique samples out of the number of all generated samples ($n$) as $n$ goes from 1 to 340,000 for all three GANs trained on the OQMD, MP, and ICSD datasets respectively (Fig.5 (a)). First, it can be found that with the generation of more and more samples, the percentage of unique samples goes down, showing that it is more difficult to generate new hypothetical materials. However, even after generating 340,000 samples, our GANs still maintain a 68.09%, 85.90%, and 73.06% uniqueness for GAN-OQMD, GAN-MP, and GAN-ICSD respectively. While all three curves decay with increasing number of generated samples, the GAN-MP maintains higher percentage of unique samples. Actually, the uniqueness curve of GAN-MP dominates the one of GAN-ICSD, which further dominates the one of GAN-OQMD. After close examination of the distributions of training and generated samples in terms of their element numbers, we found that this is mainly due to the distribution bias of the training sets of the three GANs (See supplementary Fig. S3). For GAN-OQMD, the training set is dominated by ternary compounds (84.4%) and it tends to generated ternary samples while the total number of chemically valid materials as estimated by SMACT(1) (Semiconducting Materials from Analogy and Chemical Theory) to be around 200,000. So, it tends to generate many duplicate ternary samples. For GAN-ICSD, the ratio of binary/ternary/quaternary is about 2:3:1, which allows it to generate more diverse samples, leading to higher uniqueness curve. For GAN-MP, the ratio of binary/ternary/quaternary is about 0.8:2:1, which is much more balanced than those of GAN-OQMD and GAN-ICSD and it also has much more quaternary and quinary training samples (See supplementary Table S4). This allows it to generate most diverse samples.

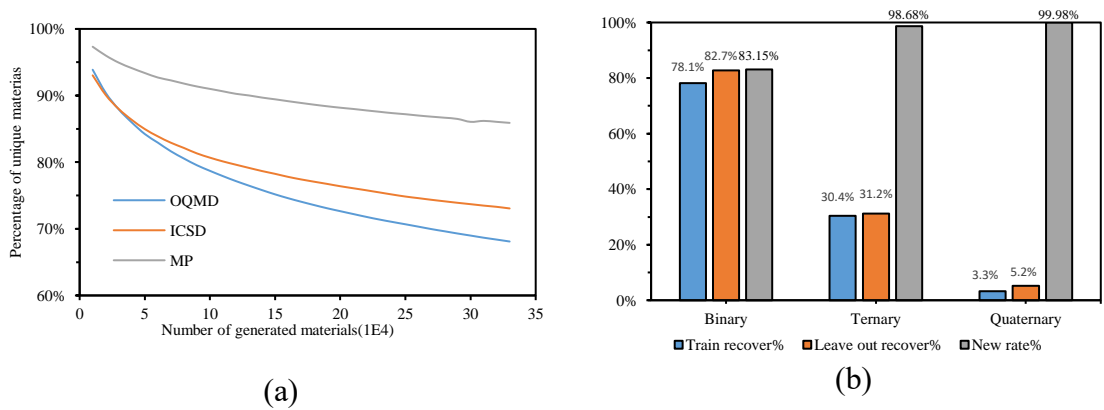

(a) (b)

Fig. 5. Uniqueness and novelty check of the generated materials. (a) Comparison of uniqueness curves of the hypothetical materials generated by three GANs. GAN-MP achieves the dominating curve due to its more balanced distribution of binary/ternary/quaternary training samples. (b) Distribution of recovery rates of training and validation samples and also percentages of new generated hypothetical materials.

**Novelty check**: to check the capability of our GANs to generate novel materials, we use the hold-out validation approach. We first leave out 10% samples from each of the three datasets OQMD, MP, and ICSD. Then we train the GANs and use them to generate a certain number of samples. We then examine what percentage of training samples and hold-out validation samples have been recovered/re-discovered and how many new samples have been generated. The results are shown in Table 1. First we found that when the GANs recover/generate a certain percentage of training samples, the approximate corresponding percentages of validation (hold-out novel) samples are also recovered. For example, when the GAN-MP recovered 47.36% of its training set, about 48.82% of the hold-out samples have also been simultaneously generated. This demonstrates that our GANs can be used to discover new materials that do not exist in the training set. To further understand the generation performance, we calculated the recovery percentages of the training set and the leave-out validation set along with the percentages of new samples for binary, ternary, and quaternary samples (Fig. 5(b)). First, by generating 2 million samples, GAN-ICSD has generated 78.1% training binary samples while also generating/rediscovering 82.7% leave-out validation binary materials. The recovery rates drop to 30.4% and 31.2% respectively or ternary training and validation samples as the number of possible ternary samples are larger than binary ones, which also explains the recovery rates dropping to 3.3% and 5.2% for quaternary training and validation sets. In addition, out of all the generated binary/ternary/quaternary samples, 83.15%/98.68%/99.98% of them are novel hypothetical materials, which strongly shows the capability of our GAN model to generate new materials candidates as a majority of these new candidates satisfy the basic chemical rules as shown in Fig.4.

Table 1 Novelty check of generated samples by GANs.

|  | GAN-OQMD | GAN-MP | GAN-ICSD |
| --- | --- | --- | --- |
| Training sample # | 251,368 | 57,530 | 25,323 |
| Leave out sample # | 27,929 | 6,392 | 2,813 |
| Generated sample # | 2,000,000 | 2,000,000 | 2,000,000 |
| Recovery % of training samples | 60.26% | 47.36% | 59.54% |
| Recovery % of leave out sample | 60.43% | 48.82% | 60.13% |
| New samples | 1,831,648 | 1,969,633 | 1,983,231 |

**Conditional generation of hypothetical materials by GAN:** in addition to generating valid inorganic materials, it is interesting to check if our GAN models can generate new materials with desired properties by sampling from the generative distribution estimated by the model (25). To verify this, we collected 30186 inorganic materials from Materials

Project whose band gap values are larger than 0. We then use these high-bandgap materials set to train a GAN-Bandgap model aiming to generate hypothetical high-bandgap materials. To verify the band gap values of generated samples, we trained a bandgap prediction model using the Gradient Boosted Decision Tree (GBDT) machine learning algorithm with Magpie features (26) (See Methods part for its training details). We also use this model to predict the bandgap values of the exhaustively enumerated materials set. Fig. 6 shows the distribution of the band gaps of the generated materials set versus those of the training set and the exhaustively enumerated set. The bandgap distribution of generated samples is much similar to that of the training set, which demonstrates the capability of our GAN-bandgap can generate hypothetic high-bandgap materials efficiently.

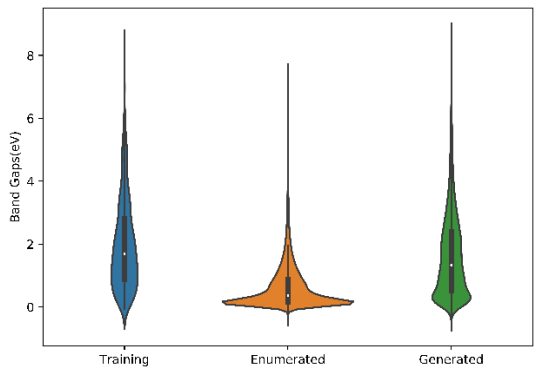

Fig. 6. Comparison of bandgap distributions of the generated materials by GAN-Bandgap, the training set, and the enumerated set

**Discovery of potential new materials:** To evaluate how likely our GAN models can generate confirmed new materials, we take a cross-validation evaluation approach. Essentially, for all the new hypothetical materials generated by each of our GAN models, we check how many of them are confirmed/included by the other two data sets. Table 2 lists the cross-validation confirmation results. It is found that out of the 2 million generated materials by GAN-ICSD, 13,126 materials are confirmed by and included in the MP dataset and 2,349 new materials are confirmed by the OQMD dataset. GAN-MP also has 6,880 and 3,601 generated samples confirmed by ICSD and OQMD, respectively.

Table 2. Cross-validation confirmation of generated new materials by our GANs

|          | ICSD dataset | MP dataset | OQMD dataset |
|----------|--------------|------------|--------------|
| GAN-ICSD | N/A          | 13,126     | 23,49        |
| GAN-MP   | 6,880        | N/A        | 3,601        |
| GAN-OQMD | 3,428        | 58,603     | N/A          |

**Limitation of MatGAN examined by Autoencoder:** Here we aim to check the relation of AE non-decodable materials and the difficulty of our GANs to generate them. To train the AE model, we randomly split the OQMD_L dataset with 90% samples for AE training and 10% samples as testing. The learning rate is set as $10^{-3}$, batch size 1024, and Adam optimizer is used. The final AE model is picked as the model with the best performance over the test set within 1000 epochs of training. We found that our AE model can decode 96.31% and 95.50% of the samples from the training set and the test set. These samples seem to share some common chemical composition rules.

To show the difference between the decodable samples and non-decodable ones, we applied T-sne dimension reduction technique (23) to reduce the dimension of the matrix representations of all OQMD_L dataset to 2 and then visualize 20% of the samples on the 2D plot (Fig.10), in which the red dots represent non-decodable samples while blue ones represent decodable ones. The apparent different distributions show that these two categories of samples have different composition rules. Our hypothesis is that the decodable samples share well-established chemical composition rules, which allows our GAN generators for efficient sampling of the corresponding chemical space. On the other hand, the non-decodable samples will be difficult to generate by our GAN model. To verify this, we calculated the percentage of non-decodable samples that have been generated by the trained GAN-OQMD. It is observed that almost 95% of the non-decodable materials are out of the scope of the generated samples even after generating 2 million of samples while 60.26% of the decodable training samples have been re-discovered.

This shows that our GANs have limitation in generating non-decodable materials type. It also means that non-decodable materials have special composition rules that either need more data or more powerful generator models to learn. Indeed, comparison on the enriched element distribution analysis (Supplementary Fig. S4) shows that the decodable and non-decodable materials have distinct element distributions.

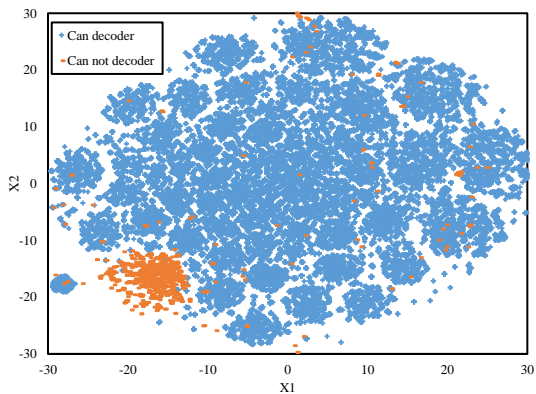

Fig.7 Distribution of decodable and non-decodable materials. X1 and X2 are the two dimensions after dimension reduction.

## 3. Discussion

The configurational phase space for new inorganic materials is immense. Forming four-component compounds from the first 103 elements of the periodic table results in more than $10^{12}$ combinations. Such a vast materials design space is intractable to high-throughput experiments or first-principle computations. On the other hand, current inorganic materials databases such as ICSD and Materials Projects all consist of only a tiny portion of the whole inorganic chemical space, which needs expansion for computational screening of new materials.

Here we proposed a GAN based generative model for efficient sampling of the vast chemical design space of inorganic materials. Systematic experiments and validations show that our GAN models can achieve high uniqueness, validity, and diversity in terms of its generation capability. Our generative models can be used to explore the uncharted inorganic materials design space by expanding ICDS, materials projects (MP), and OQMD databases. The derived expanded databases can then be used for high-throughput computational screening with higher efficiency than exhaustively screening billions of candidates (2). While principles of charge neutrality and electronegativity balance(1) have been applied to filter out chemically implausible compositions for more effective search of new materials, such explicit composition rules are still too loose to ensure efficient sampling in the vast chemical design space for new materials search. Indeed, while the hypothetical materials with less than 5 elements can be enumerated (32 billion for 4-element materials with charge-neutrality and balanced electronegativity), the design space of more elements can be challenging for which our GAN models can help a lot.

Our work can be extended in multiple ways. First, we found that MatGAN can learn chemical composition rules implicitly even though we did not explicitly enforce those rules into our GAN model. However, it is sometimes desirable to implement chemical rule filters to remove chemically invalid candidates, which can be easily implemented based on our matrix representation of materials. Another limitation in our current study is that we only considered the integer ratios of elements in compounds in our material representation while doped materials with fractional ratios are very common in functional materials such as lithium ion battery material $LiZn_{0.01}Fe_{0.99}PO_4$, which is a doped cathode material. Our study can be extended by allowing real numbers on the representation matrix. However, considering the infinite possibility of doping ratios, our GAN method may need to work together with other sampling techniques such as genetic algorithms (27, 28), genetic programming (29), and active machine learning for mixed parameter search (30, 31) or the Bayesian optimization approach (30). In addition, our current GAN models do not tell the crystal structures (lattice constants, space group, atomic coordinates, etc.) of the hypothesized materials. However, with sufficient computational resources, it is possible to exploit DFT-based computational software packages such as USPEX (32) or CALYPSO (33) to determine the crystal structure given a material composition and its stoichiometry. Our GAN models can also be used to work together with material structure generators (11).

## 4. Methods
### 4.1 Datasets

We use a subset of inorganic materials deposited in the OQMD(13, 15) database to train our AE and GAN models. OQMD is a widely used DFT database with crystal structures either calculated from high-throughput DFT or obtained from the ICSD(34) database. Currently it has 606,115 compounds. We use a similar screening criteria by Jha et.al. (35) to choose the OQMD subset for GAN training: for a formula with multiple reported formation energies, we keep the lowest one to select the most stable compound. Single-element compounds are all removed along with materials whose formation energy is out of the range of {u-5σ, u+5σ}, where u and σ are the average and standard deviation of the formation energies of all samples in OQMD. Our final dataset, OQMD_L, has 291,884 compounds.

As comparison, we also train two GANs for the Materials Projects (MP) and ICSD databases respectively. Both the MP dataset and ICSD dataset here are prepared by removing all the single-atom compounds, the compounds which has any element with more than 8 atoms in their unit cell, and compounds containing Kr and He elements. The final MP dataset used here has 63,922 compounds. The final ICSD dataset used here has 28,137 compounds.

### 4.2 GAN Neural networks training

We have optimized the hyper-parameters for training the GANs by setting the learning rate from 0.1 to $10^{-6}$ (each time decrease by 10 fold) and batch normalization size from 32 to 1024, and using different optimizers. We train our GANs using the screened samples from the OQMD, MP, and ICSD, and ICSD_filter database, which is an ICSD subset with all charge-neutral and electronegativity balanced materials. These Wasserstein GANs are trained for 1000 epochs with the Adam optimizer with learning rate of 0.001 for the generator training and 0.01 for the discriminator training. The batch size for GAN training on OQMD is set to 512 while the batch sizes are set as 32 for GAN training on all other datasets. The AE is trained with the Adam optimization algorithm with a learning rate of $10^{-3}$ and batch size of 1024.

### 4.3 Training of band gap prediction model

We choose 30,186 inorganic materials whose band gaps are greater than 0 out of the 63,922 compounds in the selected MP dataset as the training samples to train the band gap prediction model. Gradient Boosted Decision Tree (GBDT) machine learning model is then trained with the Magpie features. The learning rate is set as 0.06. The maximum tree depth is set as 20. The subsample is set to 0.4. The number of estimator is set to 100.

Supplementary video (published on Oct. 22, 2019)

https://www.youtube.com/watch?v=psneoau1m-8

**Acknowledgements**

This work was partially supported by the National Natural Science Foundation of China under grant no. 51741101. J.H., Y.Z. and M. H. also acknowledge the support from the National Science Foundation with grant no. 1905775, 1940099, and OIA-1655740. S.L. is partially supported by National Major Scientific and Technological Special Project of China under grant no: 2018AAA0101803 and also by Guizhou Province Science & Technology Plan Talent Program # [2017]5788. We would like to thank Zhuo Cao, Chengcheng Niu, Rongzhi Dong for helpful discussions.

**Author contribution:**

J.H. conceived the project. J.H. and Y.D. developed the methodology. Y.D. implemented the method. Y.D. and Y. Z. performed the calculations. Y.D., X.L., and Y.Z. prepared the figures. J.H., Y.D., X.L. and M.H. interpreted the results. J.H and Y.D. wrote the manuscript. J.H. and S.L. supervised the project. All authors reviewed and commented on the manuscript.